\documentclass[pmlr,twocolumn,10pt]{jmlr} 
\usepackage{booktabs}
\usepackage{tabulary}
\usepackage{float}
\usepackage[load-configurations=version-1]{siunitx} 
\newcommand{\qLST}{\textit{qLST }}
\newcommand{\comment}[1]{}
\newcommand{\equal}[1]{{\hypersetup{linkcolor=black}\thanks{#1}}}

\usepackage{ifthen}

\theorembodyfont{\upshape}
\theoremheaderfont{\scshape}
\theorempostheader{:}
\theoremsep{\newline}

\jmlrvolume{LEAVE UNSET}
\jmlryear{2021}
\jmlrsubmitted{LEAVE UNSET}
\jmlrpublished{LEAVE UNSET}
\jmlrworkshop{Machine Learning for Health (ML4H) 2021} 

\title[qLST]{
\centering Interpretable ECG classification via
a query-based latent space traversal (\textit{qLST})}

\author{%
\Name{Melle B. Vessies}\equal{These authors contributed equally} \Email{m.b.vessies-2@umcutrecht.nl} \\
\addr University of Amsterdam, The Netherlands \\
\addr Department of Cardiology, University Medical Center Utrecht, Utrecht, the Netherlands
\AND
\Name{Sharvaree P. Vadgama}\footnotemark[1] \Email{s.p.vadgama@uva.nl} \\
\addr University of Amsterdam, The Netherlands
\AND
\Name{Rutger R. van de Leur} \Email{r.r.vandeleur@umcutrecht.nl} \\
\addr Department of Cardiology, University Medical Center Utrecht, Utrecht, the Netherlands
\AND
\Name{Pieter A. Doevendans} \Email{p.doevendans@umcutrecht.nl} \\
\addr Department of Cardiology, University Medical Center Utrecht, Utrecht, the Netherlands
\AND
\Name{Rutger J. Hassink} \Email{r.j.hassink@umcutrecht.nl} \\
\addr Department of Cardiology, University Medical Center Utrecht, Utrecht, the Netherlands
\AND
\Name{Erik Bekkers}
\Email{e.j.bekkers@uva.nl} \\
\addr University of Amsterdam, The Netherlands
\AND
\Name{René van Es}
\Email{r.vanes-2@umcutrecht.nl}\\
\addr Department of Cardiology, University Medical Center Utrecht, Utrecht, the Netherlands
}

\begin{document}

\maketitle

\begin{abstract}
Electrocardiography (ECG) is an effective and non-invasive diagnostic tool that measures the electrical activity of the heart. Interpretation of ECG signals to detect various abnormalities is a challenging task that requires expertise. Recently, the use of deep neural networks for ECG classification to aid medical practitioners has become popular, but their black box nature hampers clinical implementation. Several saliency-based interpretability techniques have been proposed, but they only indicate the location of important features and not the actual features. 
We present a novel interpretability technique called \textit{qLST}, a query-based latent space traversal technique that is able to provide explanations for any ECG classification model. With \textit{qLST}, we train a neural network that learns to traverse in the latent space of a variational autoencoder trained on a large university hospital dataset with over 800,000 ECGs annotated for 28 diseases. We demonstrate through experiments that we can explain different black box classifiers by generating ECGs through these traversals.
\end{abstract}

\section{Introduction}
\label{sec:intro}
Deep learning methods are increasingly being applied to tasks in the medical domain and show remarkable performance. Despite these promising developments, only few algorithms have been applied in clinical practice. The lack of interpretability of these models makes it difficult to detect biases or inaccuracies, and gain physicians' trust \citep{RudinNatMed}. These issues have also been acknowledged by the European Union's General Data Protection Regulation, that mandates a 'right to explanation' for every artificial intelligence (AI) algorithm \citep{GoodmanAI}.

The electrocardiogram (ECG) is one of the most fundamental tools in clinical practice and automated diagnosis of ECGs could be of great support in clinical practice when expert knowledge is not readily available. Therefore, a variety of deep neural networks (DNNs) have recently been developed to classify ECGs \citep{hong2020, Hannun2019}. Some of the previous studies have tried to provide interpretability of their models, but these approaches were predominantly saliency-based \citep{leur2020, leur2021, kwon2020}. Saliency-based methods only point to a location in the ECG, but do not show what the important feature at that location is. Moreover, \cite{adebayo2018, adebayo2018-2} describe certain issues concerning the trustworthiness of saliency-based explanation methods.

Recent work by \cite{vansteenkiste2019generating} and \cite{bos2020} introduced traversals of single variables in the latent space of a variational auto-encoder (VAE) as a way of visualizing meaningful ECG features. These interpretable latent space variables are subsequently used in common statistical methods, such as logistic regression. However, such methods are only used to visualize traversals of single latent space variables and do not provide a way to explain black-box classifiers. \citet{singla2020explanation} proposed a query based method to generate explanations of a black-box classifier. Their method relies on progressively making small perturbations to an input that change the posterior class probability and can only provide contrastive explanations. 

In this paper we propose a new query-based latent space traversal (\textit{qLST}) method that learns to make perturbations in the latent space of a VAE, allowing us to generate informative ECG samples in a flexible way that is not limited to contrastive explanations. 

\section{Method}
Given a classifier that maps ECGs to a set of labels, \qLST is able to generate ECGs that explain the classification process. To generate these explanations two additional models are trained, a VAE, to model the latent space and \qLST, to traverse that latent space. \qLST consists of a single attention module that learns to change multiple latent variables at once to show variation \textit{within} and \textit{between} classes. We query the latent space such that we increase the posterior probability of a class by increasing the query value. In order to \textit{see} these changes an ECG is generated using a decoder and visualized. 

\subsection{Querying}\label{sec:querying}
The querying part of the method depends on a classifier that takes an input $x$ and returns a classification label $\hat{y}$ that approximates the true class label $y$ and an encoder of a VAE that takes a raw ECG as input and outputs a latent representation ($z$). \qLST takes this latent representation along with a query value $q$ and returns a $\Delta z$, which approximates the required step in the latent space to move the posterior class probability of $x$ towards $q$ (equation \ref{eq:LST_main}).

\begin{equation}
    \centering
    \Delta z = qLST(z, q)
\label{eq:LST_main}
\end{equation}

Through querying we approximate the new classifier output label $y_{LST}$ from the initial $\hat{y}$. Higher query values correspond to higher posterior probability for a class. By generating $\Delta z$ we can traverse the latent space such that the reconstructed ECGs can traverse the label space via equation \ref{eq:lst_pipeline} . 

\begin{equation}
    \centering
    y_{LST} = (\textit{Classifier}(\textit{Decoder}(z + \Delta z))
\label{eq:lst_pipeline}
\end{equation}

\begin{figure}
    \centering
    \includegraphics[width= .35
    \textwidth]{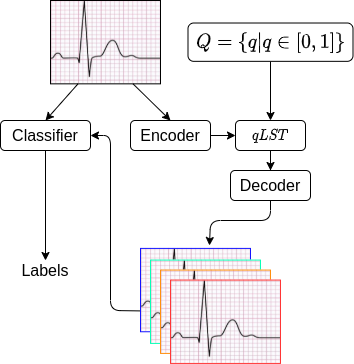}
    \caption{The \qLST takes the encoded ECGs and a set of queries ($Q = \{q|q\in[0,1]\}$) as input and \textit{generates} ECGs which are then classified. The generated ECGs and their classification labels help explain the decision process of the classifier.}
    \label{fig:qlst_pipeline}
\end{figure}

To generate a diverse set of insightful explanations we use a range of query values ($Q = \{q|q\in[0,1]\}$) that correspond to different posterior probabilities for a class. In our experiments this range was set to $Q=\{0, 0.2, 0.4, 0.6, 0.8, 1\}$. The full \qLST pipeline used to create explanations is shown in \ref{fig:qlst_pipeline}.

\begin{figure*}[h]
    \vspace{-0.3cm}
    \centering
    \includegraphics[width=0.9\textwidth, height=6cm]{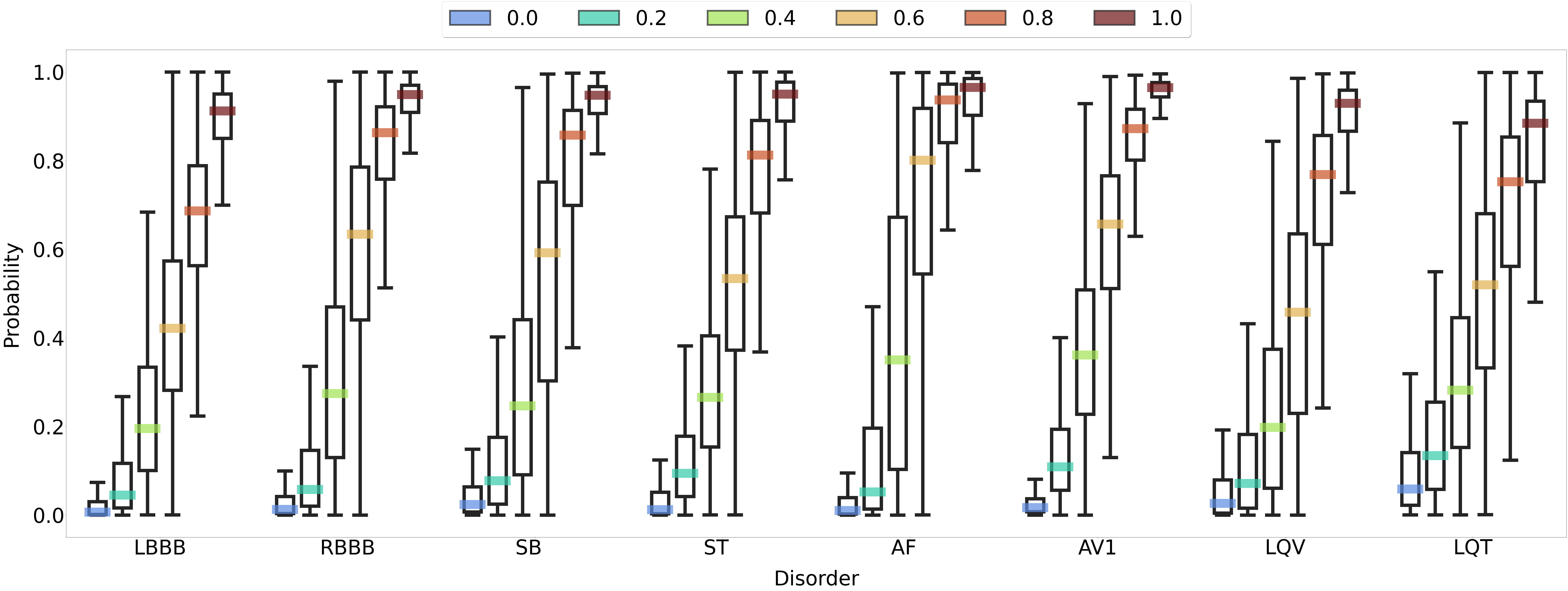}
    \caption{Relationship of the class probabilities with query values 0, 0.2, 0.4, 0.6, 0.8 and 1 for left bundle branch block (LBBB), right bundle branch block (RBBB), sinus bradycardia (SB), sinus tachycardia (ST), atrial fibrillation (AF), first degree AV block (AV1), low QRS voltage (LQV) and long QT interval (LQT). The \qLST results shown are created using the ECGResNet classifier architecture. The figure shows how \qLST is able to approximate class probabilities for the given queries.}
    \label{fig:ECGnet_boxplot}
\end{figure*}

\begin{figure*}[h]
    \vspace{-0.3cm}
  \centering
    \begin{minipage}[h]{\textwidth}
        \centering\includegraphics[width=0.9\textwidth]{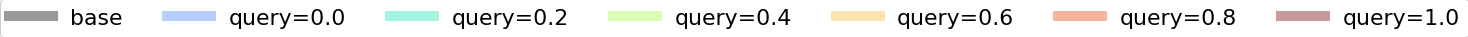}
    \end{minipage}
    \begin{minipage}[h]{\textwidth}
        (a)\includegraphics[width=0.45\textwidth]{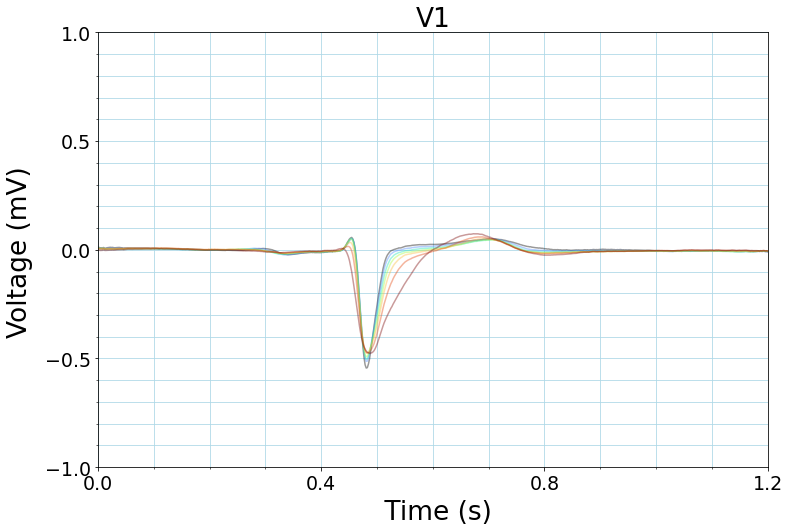}
        (b)\includegraphics[width=0.45\textwidth]{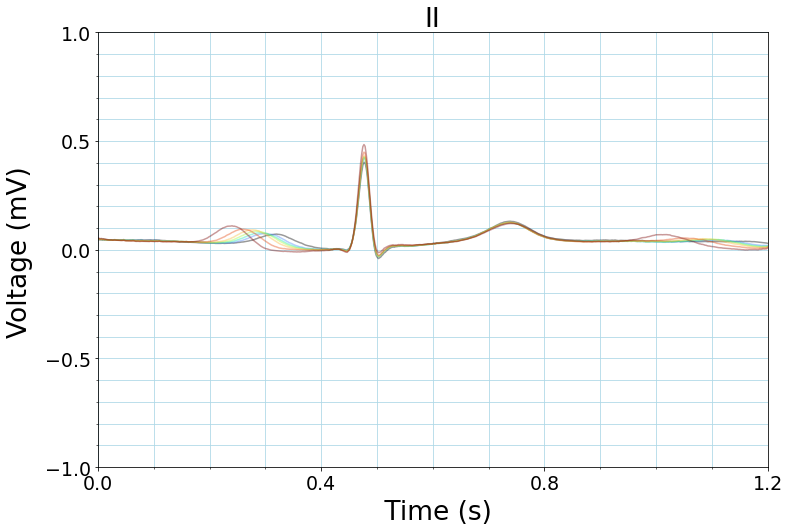}
       \caption{\textbf{(a)} Lead V1 latent traversal for a left bundle branch block (LBBB) using the qLST system. It shows the widening of the QRS complex as well as an increase in the T-wave height. Both are known features of the LBBB class. \textbf{(b)} Lead II latent traversal for a first degree AV (AV1) block using the qLST system. This shows an increase in the PR interval. A PR-interval of more than 200ms is the definition for AV1.}
        \label{fig:ecgs}
    \end{minipage}
\end{figure*}

\subsection{Training}
For the training of the $qLST$ model, different $q$ values were sampled from a uniform distribution. For the loss function, we use the binary cross entropy (BCE) loss and mean squared error (MSE) loss. The BCE is computed between the query and $y_{LST}$. To ensure that $x_{LST}$, reconstructed from $z + \Delta z$, is in fact local, the MSE loss term is used. The scaling weight $\alpha$ is chosen proportional to the difference between $q$ and the original classification label of $x$, $\hat{y}$. This enforces locality when the query values are very similar to $\hat{y}$, and $x_{LST}$ needs to be similar to $\hat{x}$ (the reconstruction of $z$ without $\Delta z$), while allowing larger changes in $x_{LST}$ only when the difference between $q$ and $\hat{y}$ becomes large. 

\comment{
\begin{equation}
    \centering
    \alpha = 25 *(1 - [q - \hat{y}])
\end{equation}
}
\begin{equation}
    \label{eq:loss}
    \centering
    L = BCE(q, y_{lst}) + \alpha MSE(\hat{x}, x_{lst})
\end{equation}

\subsection{Local and global explanations}
The \qLST model is trained on individual samples and can thus be applied to individual samples to explain a classifier locally. \qLST can explain global trends and provides interpretability by visualizing the generated ECGs for each of the query values, starting from a baseline \textit{mean} ECG where all variables in the latent space are set to zero.

\section{Experiments}
Experiments were performed with three different classifiers, a simple MLP-based classifier, ECGResNet \cite{van2020automatic} and ResNet based classifier \cite{Hannun2019} (with reduced depth). For all the experiments we use a  ControlVAE (\cite{shao2020controlvae}) with a ResNet architecture inspired by \cite{Hannun2019} to get a latent space for the dataset consisting of 864,051 ECGs from a diverse set of patients. For these ECGs we used a single median beat representation created using the MUSE system (GE Healthcare). The latent space is visualized using UMAP \citep{mcinnes2018umap}.
 
Each classifier was trained to recognize $28$ different physician annotated ECG labels. A subset of $8$ labels with clear definitions were selected to validate the proposed interpretability technique: left bundle branch block (LBBB), right bundle branch block (RBBB), sinus bradycardia (SB), sinus tachycardia (ST), atrial fibrillation (AF), first degree AV block (AV1), low QRS voltage (LQV) and long QT interval (LQT).

To evaluate the performance of $\qLST$ for local explanations, we compared different class probabilities using different query values for a given classifier. By increasing query values, the generated ECGs are expected to match the true ECG for a class (as represented by increasing class probabilities).

To confirm that \qLST is able to create global explanations, visual inspections were performed on latent traversals for the LBBB and AV1 classes starting from a generated zero latent space, which represents the mean ECG. 

The global explanations per class and local ECG explanations, along with \qLST reconstructed ECGs (for different query values), were evaluated by medical professionals to ensure that \textit{generated} ECGs showed the known features of that class and realistic generated ECGs.

\subsection{Results}

Figure \ref{fig:ECGnet_boxplot} shows the relationship of the query values with the corresponding probability for the ECGResNet model for the testset of $37,519$  ECGs (10\% of our dataset). This indicates that the \qLST can traverse through the latent space and understand the decision boundary of the classifier. Results for the other architectures are shown in the Appendix \ref{apd:first}.

Using a \qLST model trained for LBBB, we start from a latent space with all variables set to zero (representing the baseline \textit{mean} ECG) and show that the suggested traversals in latent space generate ECGs which exhibit features in agreement with the definition of an LBBB (Figure \ref{fig:ecgs}(a)). Furthermore, for the network trained for the first degree AV block class, the visualizations show that the \qLST is able to visualize an increasing PR interval (Figure \ref{fig:ecgs}(b)). The full 8-lead ECG for the classes are shown in Appendix \ref{apd:first} (Figures \ref{fig:ecg_a_full}(a) and \ref{fig:ecg_a_full}(b)). These traversals show that our network is able to recognize and show the features associated with the classes while maintaining the features of the original ECG.  

For any given ECG, \qLST provides local explanations by visualizing generated ECGs with the given ECG as a starting point. From that ECG, we are able to both add and remove a specific class and visualize the changes to the ECG. Details are shown in the Appendix \ref{apd:first}.



\section{Conclusions}
In this study, we present a novel technique to traverse through the latent space of an ECG to explain any given classifier’s decisions globally and locally. \qLST provides a method that will not only allow users to better understand classifiers but also allow for the evaluation of model bias and inaccuracies. The features used by the classifier are shown by generating ECGs, which allows for a very precise visualization of the feature morphology. 

\section{Future Work}
Further validation of the method for other classes and combinations of abnormalities is needed. Moreover, we plan on be releasing an interactive visualization tool (Figure in Appendix \ref{apd:first})) to make our method more accessible for medical professionals. We also aim to address \qLST's limitations: a separate \qLST model has to be trained for each disorder and \qLST cannot directly be used to improve a classifier architecture.

\section{Acknowledgements}
This research was (partially) funded by the Hybrid Intelligence
Center, a 10-year programme funded by the Dutch Ministry of Education, Culture and Science through the Netherlands Organisation
for Scientific Research. The research is also partially financed by the NWO research programme VENI (grant number 17290), the Netherlands Organisation for Health Research and Development (ZonMw), no. 104021004, and the Dutch Heart Foundation, no. 2019B011.

\newpage
\section{Citations and Bibliography}
\label{sec:cite}


\bibliography{jmlr-sample}

\appendix

\section{}\label{apd:first}



\subsection{UMAP projections}

In figure \ref{fig:a}, clusters are observed for the sinus rhythm, sinus tachycardia, sinus bradycardia, left bundle branch block and right bundle branch block classes. Moreover, there is visible overlap between all classes except for tachycardia and bradycardia which lay on opposing sides of the latent space. This is also physiologically plausible as these two labels are mutually exclusive and have opposite physiological characteristics (fast and slow heart rates respectively), while other labels can co-exist in one patient.

\begin{figure}[H]
    \centering
    \label{fig:a}
    \includegraphics[width= 0.49\textwidth]{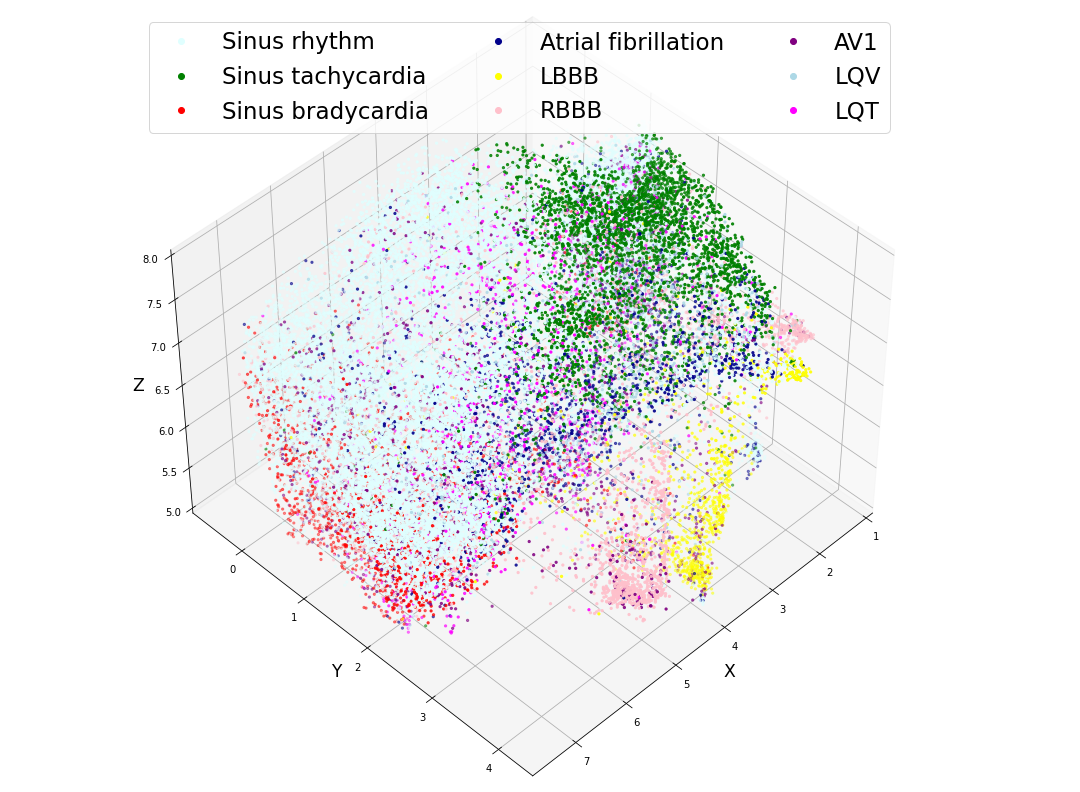}
    \caption{UMAP projection of the latent space in 3D. The visualization includes ECGs with 9 different labels: normal sinus rhythm, left bundle branch block (LBBB), right bundle branch block (RBBB), sinus bradycardia (SB), sinus tachycardia (ST), atrial fibrillation (AF), first degree AV block (AV1), low QRS voltage (LQV) and long QT interval (LQT).}
    \addtocounter{figure}{1}
\end{figure}

\subsection{Visualization tool}
\begin{figure*}[h]
    \centering
    \includegraphics[width=.9\textwidth]{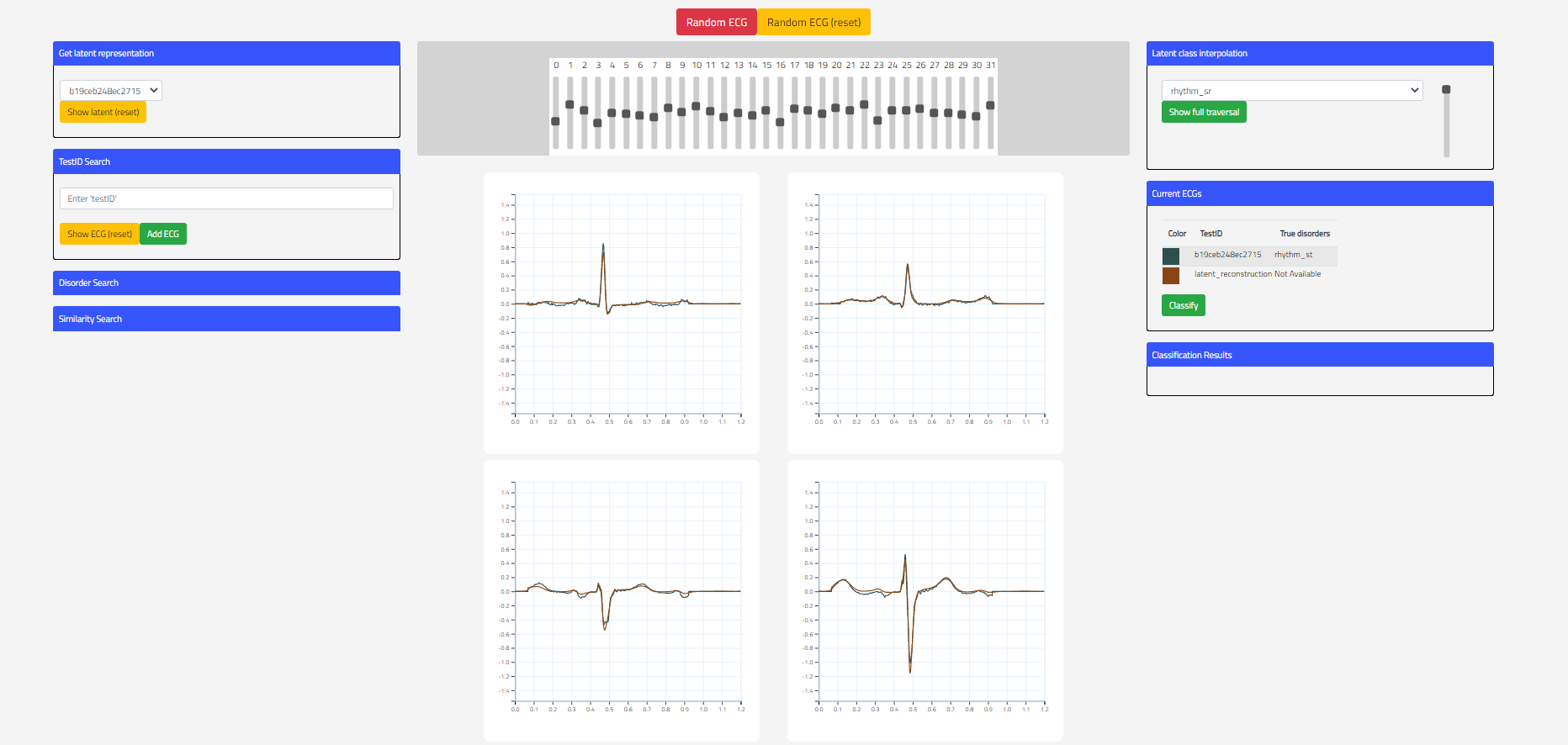}
    \caption{Figure shows a screenshot of the online visualization tool. At the top there are multiple sliders that correspond to the values of the latent variables of the VAE. Moving these sliders will change the values of the latent variables and instantly show the corresponding changes to the four channels of ECG in the window below.}
    \label{fig:screenshot}
\end{figure*}

For future work we intent to include \qLST in a interactive visualization tool for medical professionals. The tool is currently under development and a screenshot is shown in Figure \ref{fig:screenshot}. 

\subsection{All-channel ECG results}
\begin{figure*}[h]
    \vspace{-0.3cm}
  \centering
    \begin{minipage}[h]{\textwidth}
        \centering\includegraphics[width=0.9\textwidth]{images/legend.png}
    \end{minipage}
    \begin{minipage}[h]{\textwidth}
        (a)\includegraphics[width=0.45\textwidth]{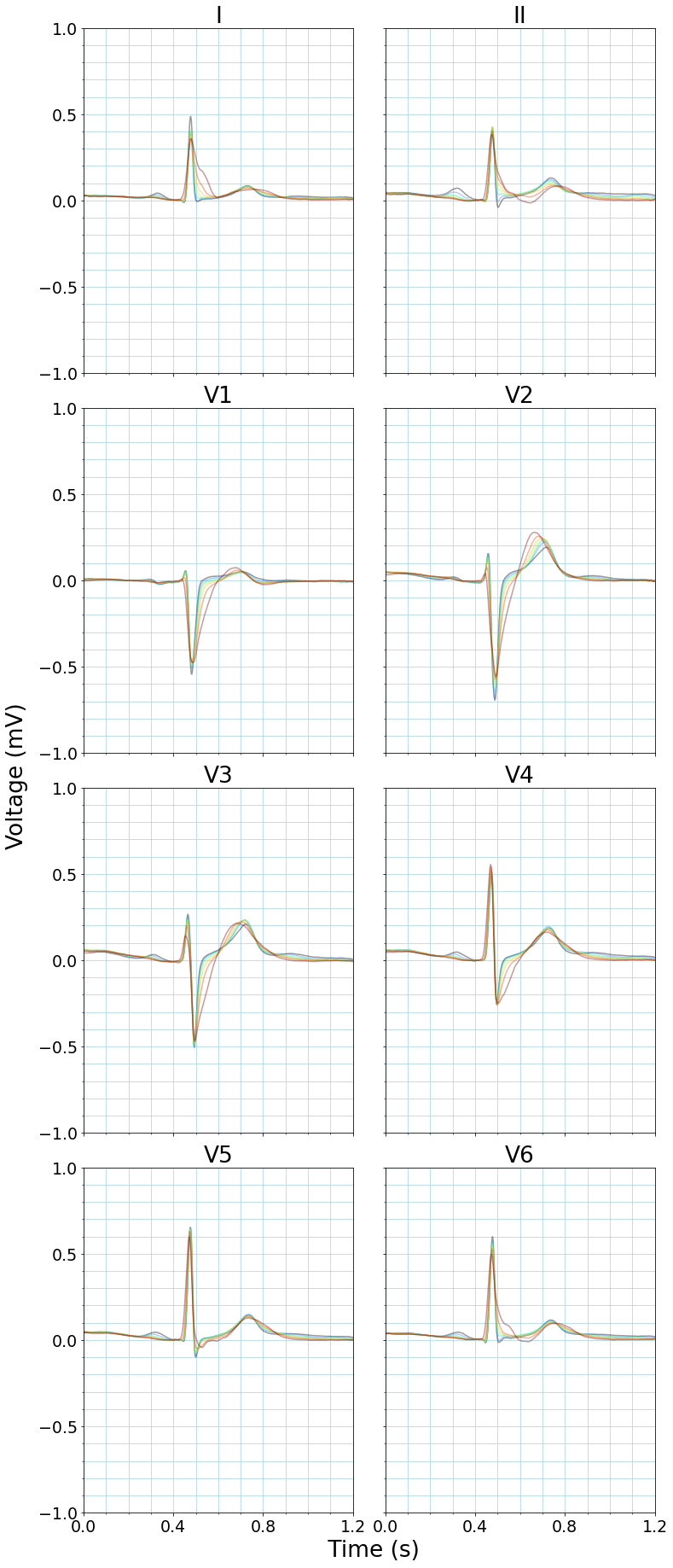}
        (b)\includegraphics[width=0.45\textwidth]{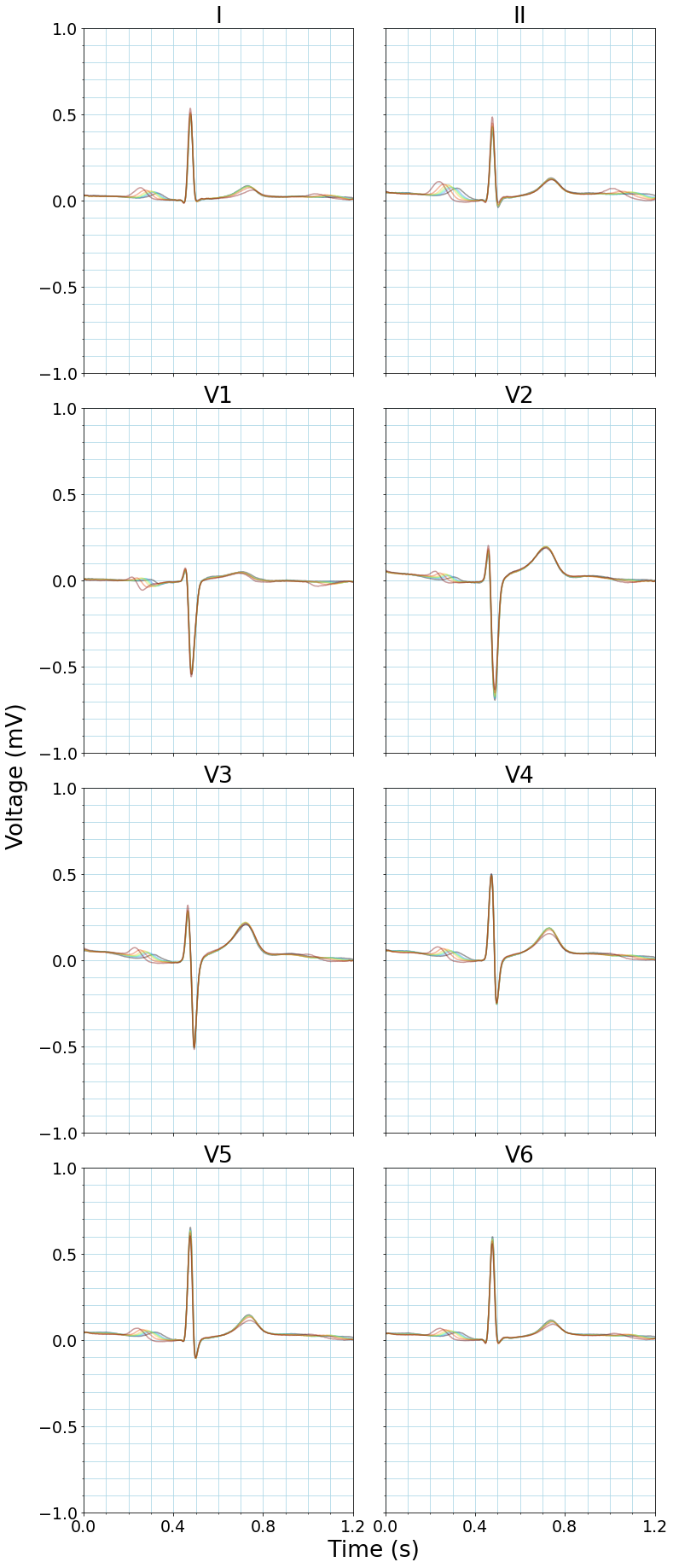}
       \caption{\textbf{(a)} 8 lead latent traversal for a left bundle branch block (lbbb) using the q-lst system. It shows the widening of the QRS complex (by widening the S- and R-waves) as well as changes in T-wave morphology's. Both these features are known features of the lbbb class. \textbf{(b)} 8 lead latent traversal for a first degree AV (av1) block using the q-lst system. This shows an increase in the PR interval. A PR-interval of more than 200ms is a defining for an av1 block}
  \label{fig:ecg_a_full}
    \end{minipage}
\end{figure*}

The figure \ref{fig:ecg_a_full}(a) shows the lbbb with different queries starting from latent variables all set to value 0. 

The figure \ref{fig:ecg_a_full}(b) shows the av1 with different queries starting from latent a real patients ECG with a normal sinus rhythm. 

\subsection{Explaining correctly classified ECG results}
\begin{figure*}[h]
    \vspace{-1cm}
  \centering
    \begin{minipage}[h]{\textwidth}
        \centering\includegraphics[width=0.9\textwidth]{images/legend.png}
    \end{minipage}
    \begin{minipage}[h]{\textwidth}
        (a)\includegraphics[width=0.45\textwidth]{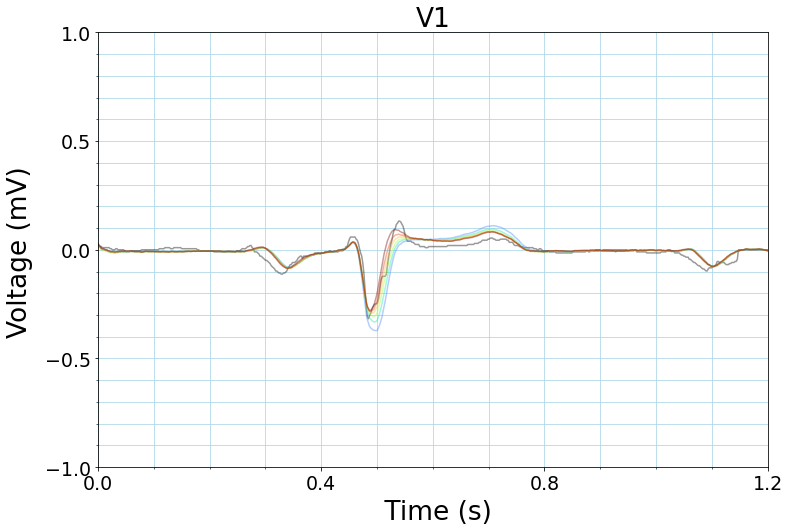}
        \hfill
        (b)\includegraphics[width=0.45\textwidth]{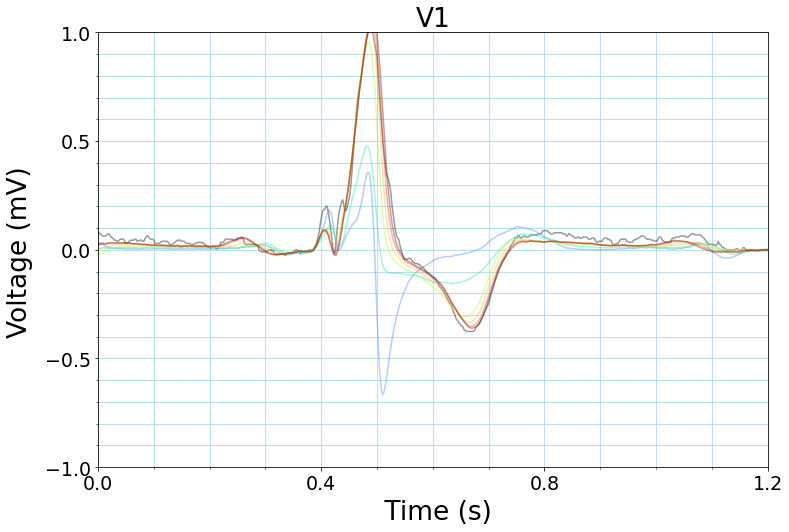}
        \hfill
        (c)\includegraphics[width=0.45\textwidth]{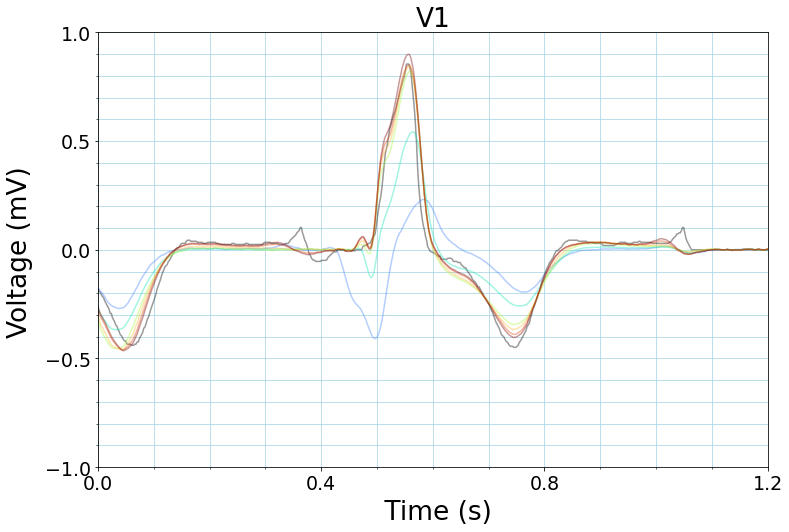}
        \hfill
        (d)\includegraphics[width=0.45\textwidth]{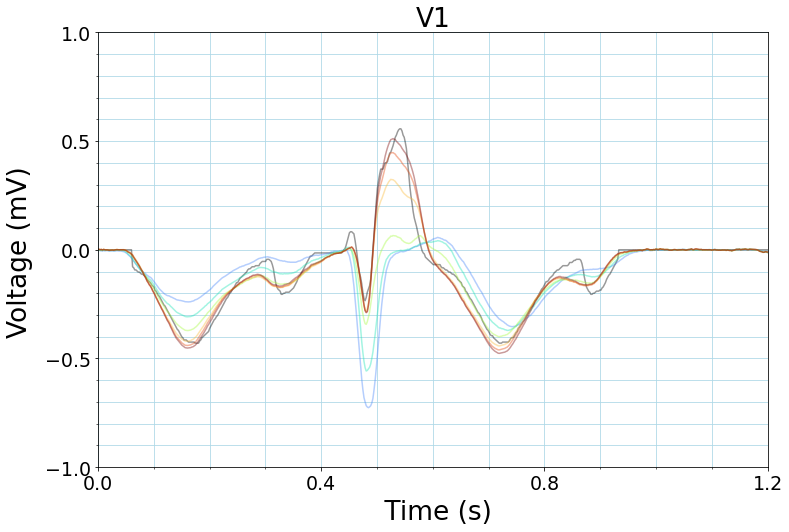}
        \hfill
        (e)\includegraphics[width=0.45\textwidth]{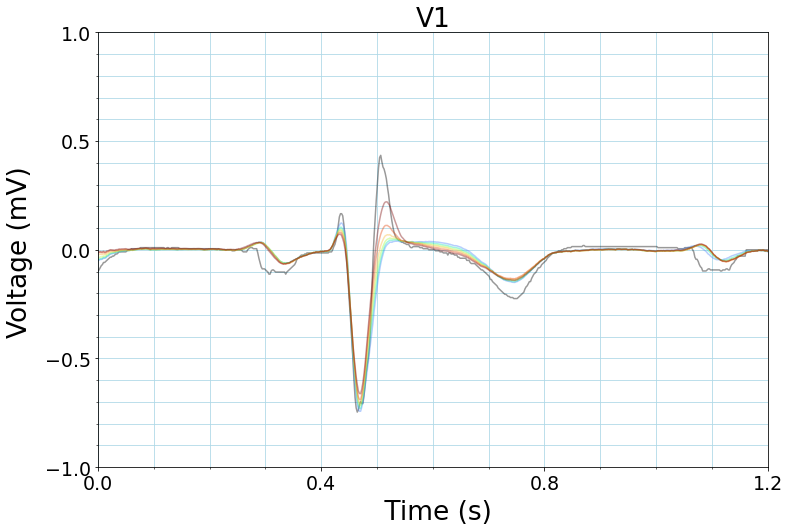}
        \hfill
        (f)\includegraphics[width=0.45\textwidth]{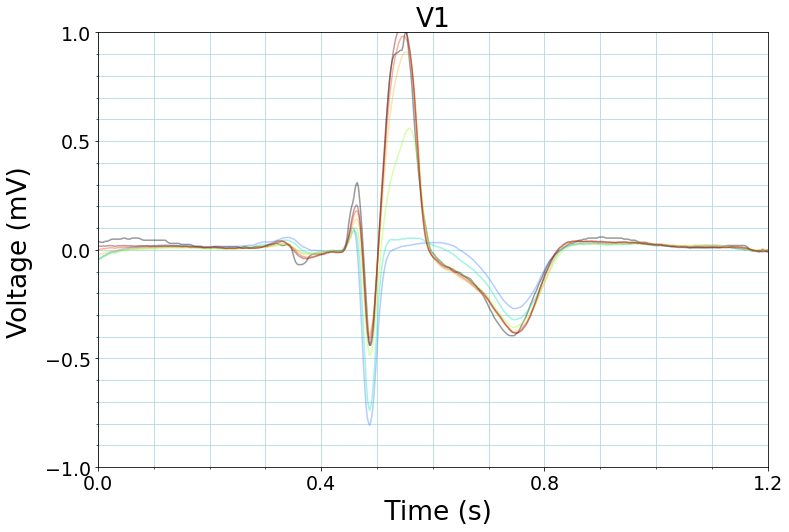}
        \hfill
        (g)\includegraphics[width=0.45\textwidth]{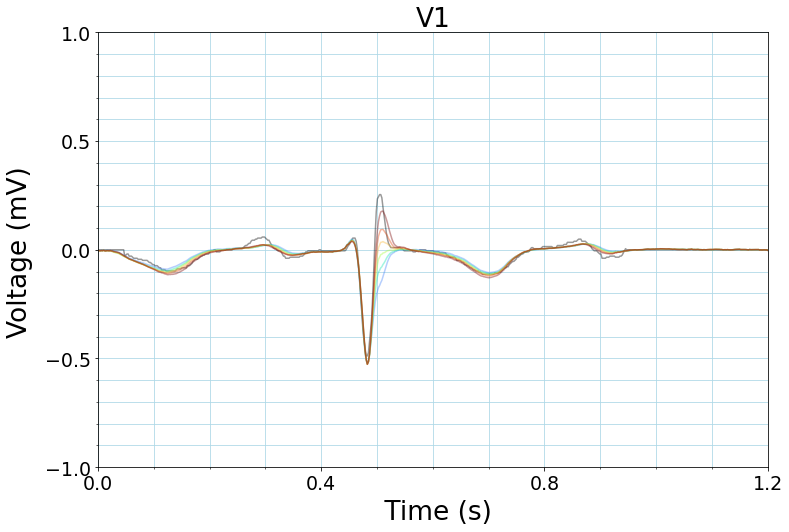}
        \hspace{0.6cm}
        (h)\includegraphics[width=0.45\textwidth]{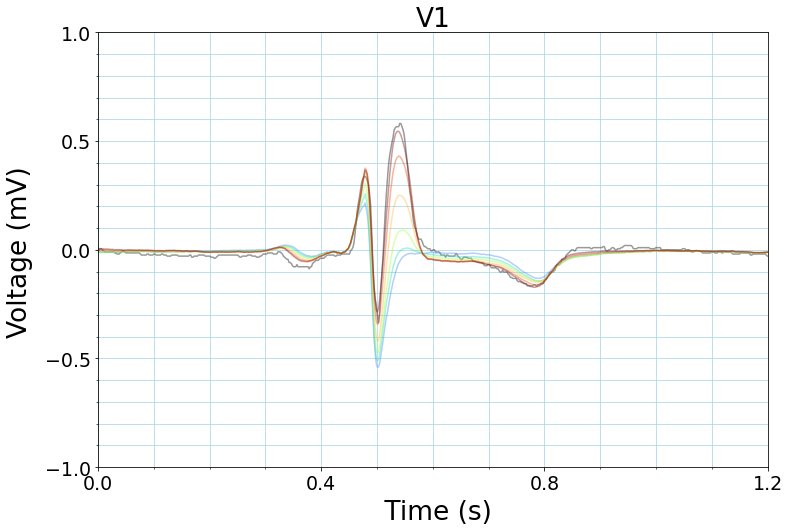}
        \hfill
        \label{fig:rbbb_reduce}
         \caption{(a) to (h) show lead V1 latent traversals using our \qLST system on right bundle
        branch block (RBBB) samples that were \textit{correctly} classified. The traversals show how \qLST is able to \textit{decrease} the posterior RBBB class probability of the samples while preserving local features.}
    \end{minipage}
\end{figure*}

\qLST can explain the classifier's decision when it has correctly classified an ECG. To do so we can reduce the posterior probability of a class and visualize the corresponding ECGs. Through \textit{qLST}, we explain correct classification in figure \ref{fig:rbbb_reduce}.

\subsection{Explaining misclassified ECG results}
\begin{figure*}[h]
    \vspace{-1cm}
  \centering
    \begin{minipage}[h]{\textwidth}
        \centering\includegraphics[width=0.9\textwidth]{images/legend.png}
    \end{minipage}
    \begin{minipage}[h]{\textwidth}
        (a)\includegraphics[width=0.45\textwidth]{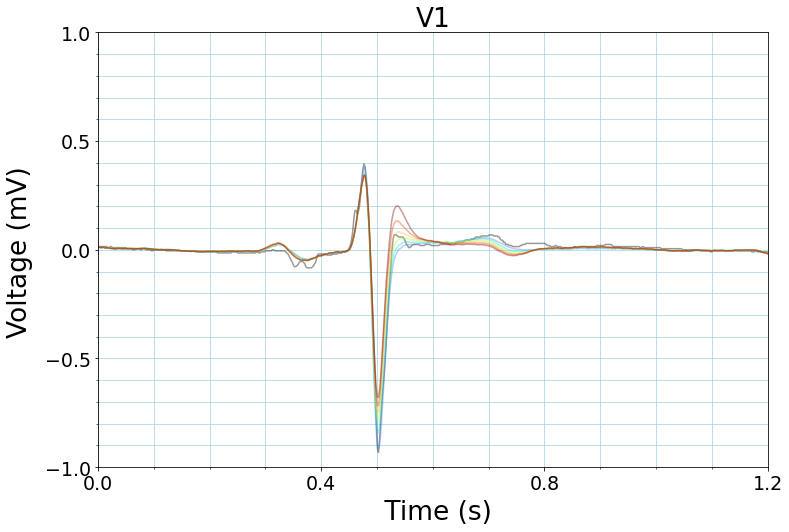}
        \hfill
        (b)\includegraphics[width=0.45\textwidth]{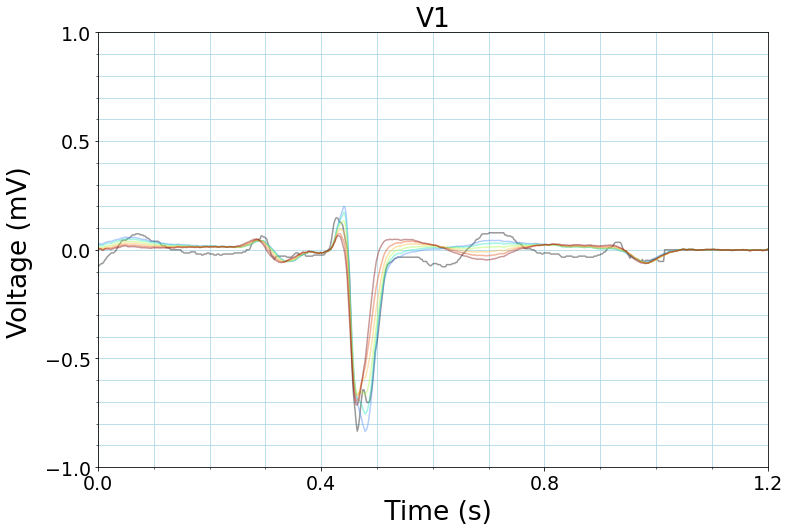}
        \hfill
        (c)\includegraphics[width=0.45\textwidth]{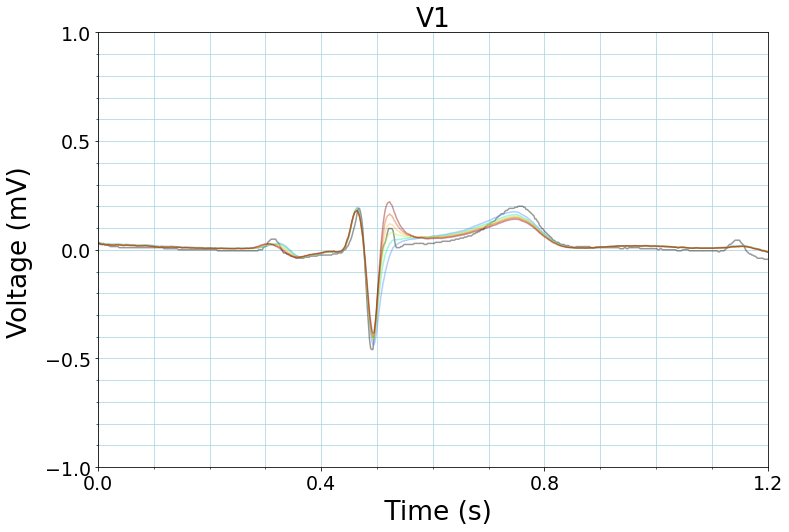}
        \hfill
        (d)\includegraphics[width=0.45\textwidth]{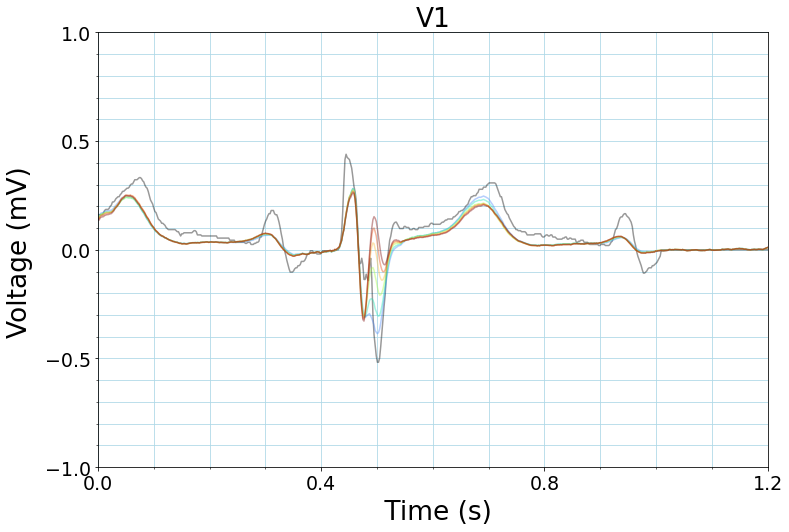}
        \hfill
        (e)\includegraphics[width=0.45\textwidth]{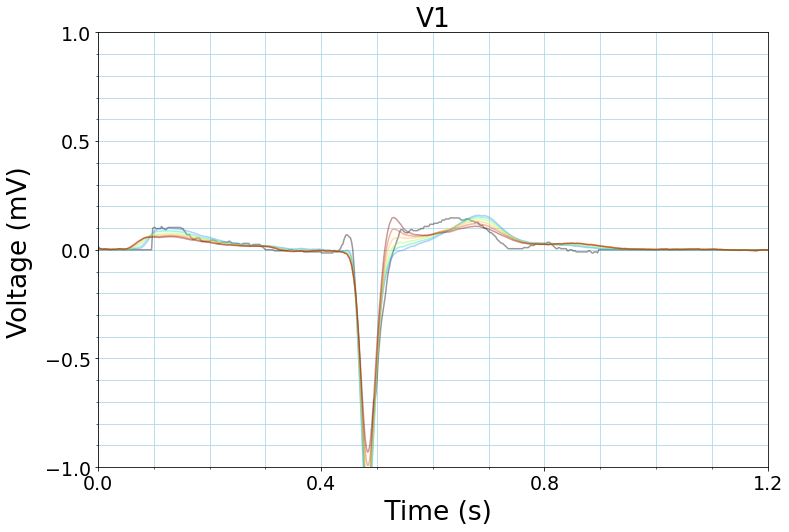}
        \hfill
        (f)\includegraphics[width=0.45\textwidth]{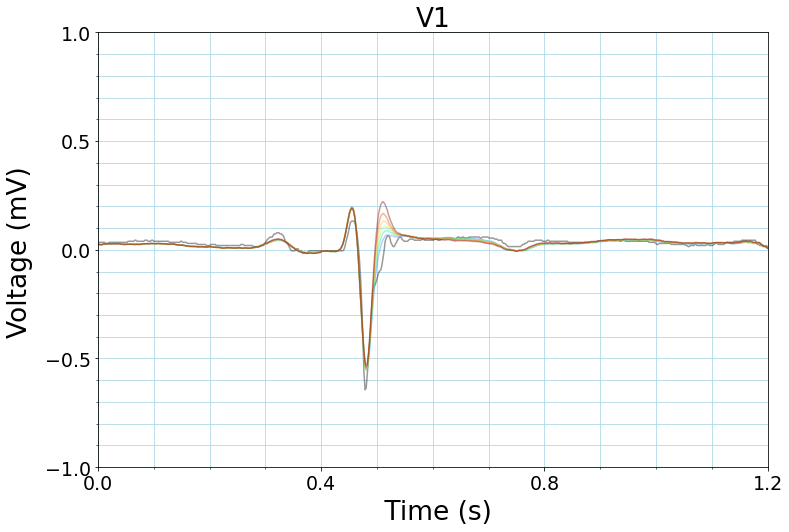}
        \hfill
        (g)\includegraphics[width=0.45\textwidth]{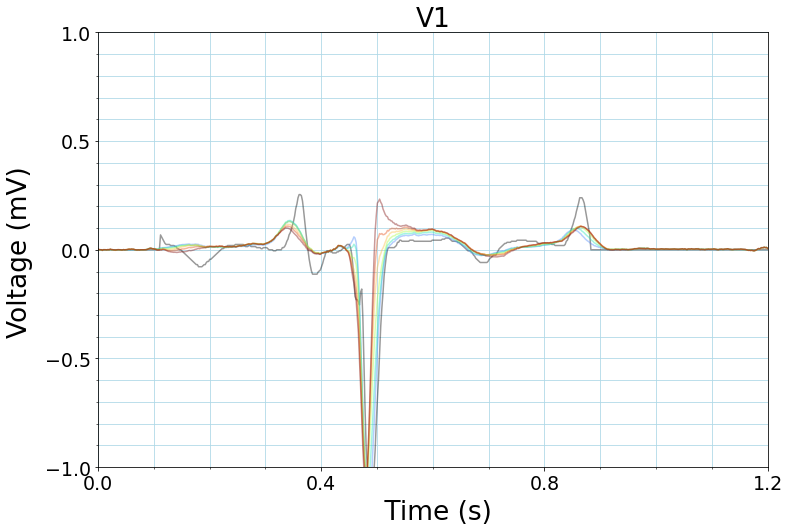}
        \hspace{0.6cm}
        (h)\includegraphics[width=0.45\textwidth]{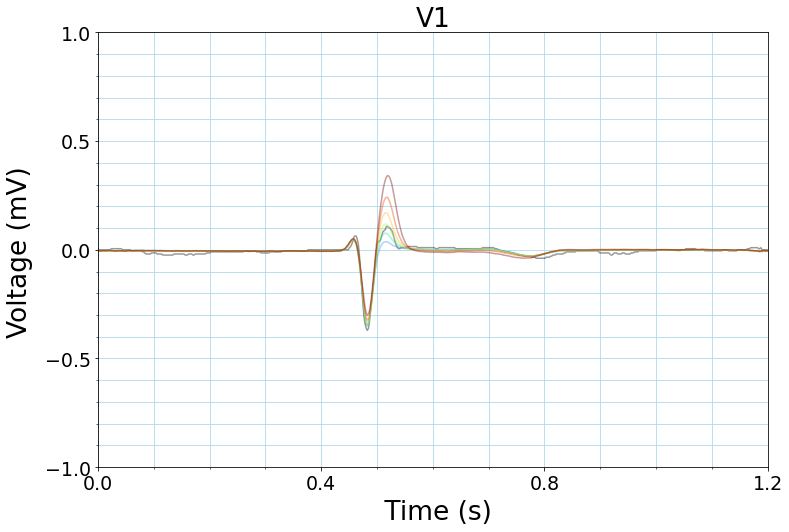}
        \hfill
       \caption{(a) to (h) show lead V1 latent traversals using our \qLST system on right bundle
        branch block (RBBB) samples that were \textit{wrongly} classified. The traversals show how \qLST is able to \textit{increase} the posterior RBBB class probability of the samples while preserving local features.}
    \label{fig:rbbb_increase}
    \end{minipage}
\end{figure*}

\qLST can explain the classifier's decision when it has misclassified an ECG. As important as it is to understand the correctly classified ECG,  understanding misclassification helps  understand induction bias and helps improve on future designs. Through \textit{qLST}, we explain misclassification in figure \ref{fig:rbtb_af}(a) and \ref{fig:rbbb_increase}.

\subsection{Successful introduction of a class}

\begin{figure*}[h]
    \vspace{-1cm}
  \centering
    \begin{minipage}[h]{\textwidth}
        \centering\includegraphics[width=0.9\textwidth]{images/legend.png}
    \end{minipage}
    \begin{minipage}[h]{\textwidth}
        (a)\includegraphics[width=0.45\textwidth]{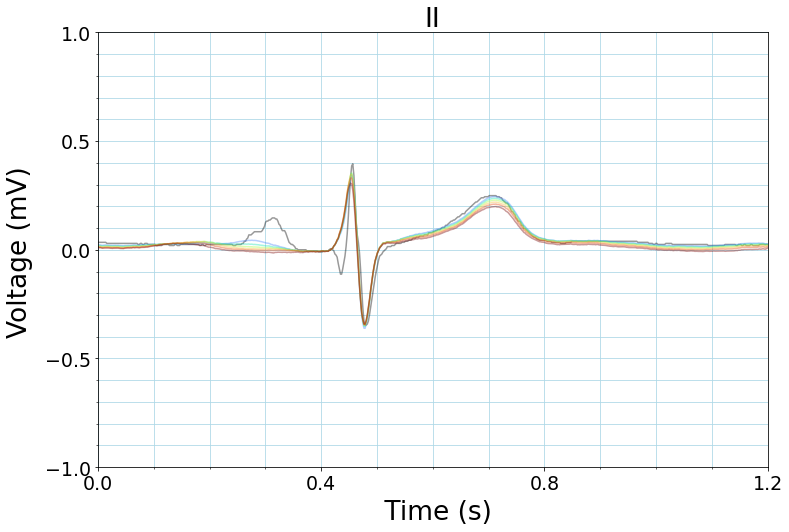}
        \hfill
        (b)\includegraphics[width=0.45\textwidth]{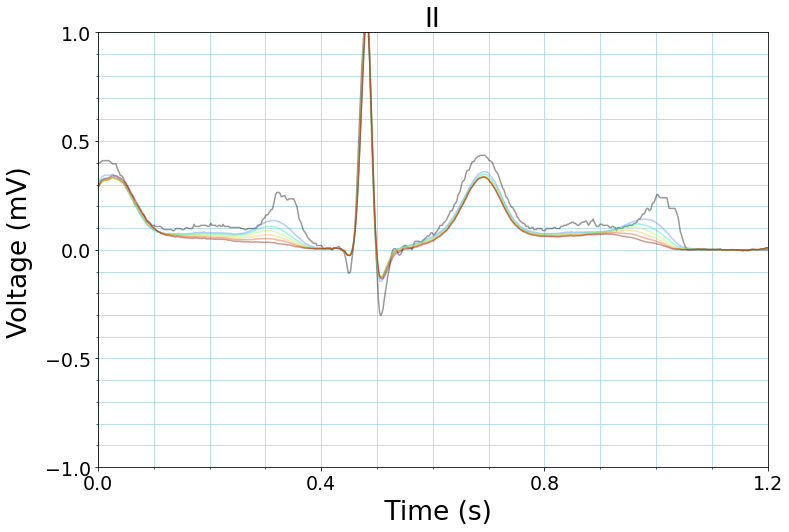}
        \hfill
        (c)\includegraphics[width=0.45\textwidth]{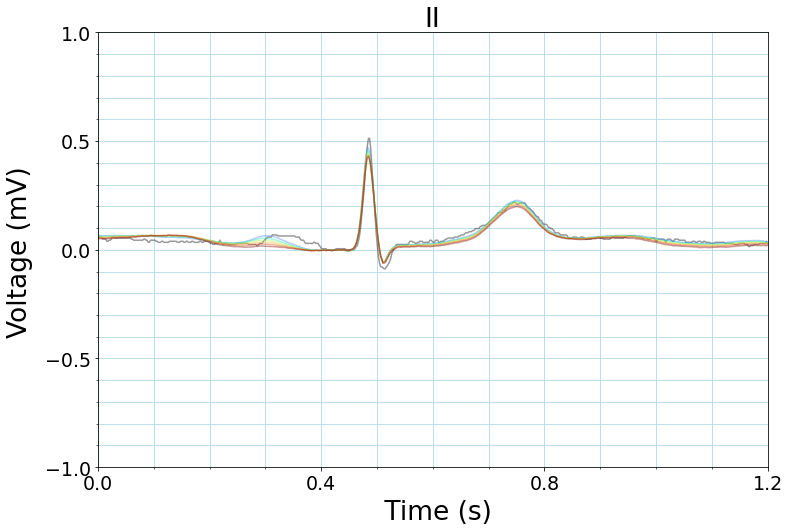}
        \hfill
        (d)\includegraphics[width=0.45\textwidth]{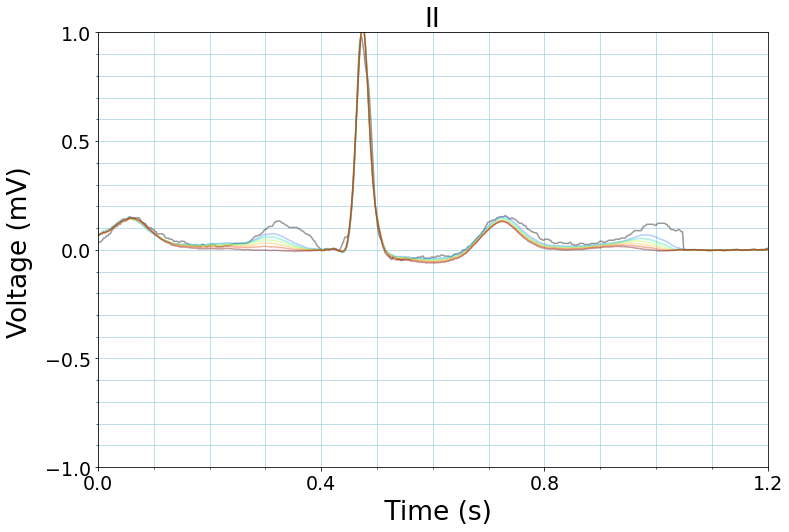}
        \hfill
        (e)\includegraphics[width=0.45\textwidth]{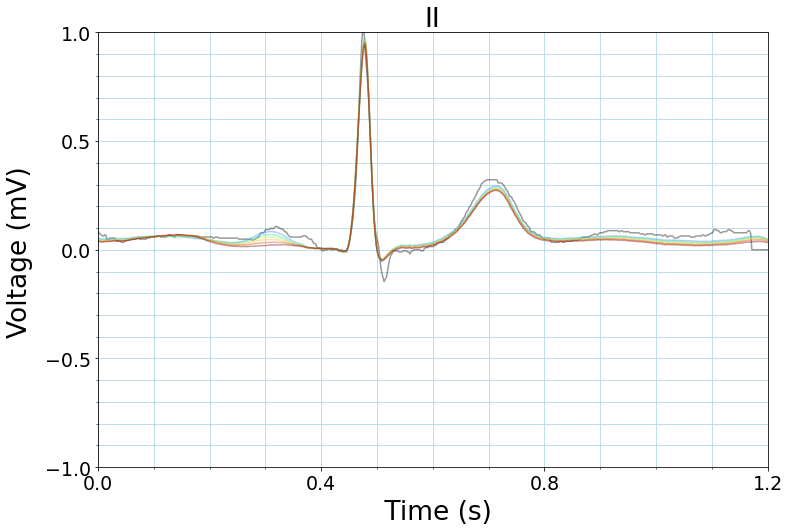}
        \hfill
        (f)\includegraphics[width=0.45\textwidth]{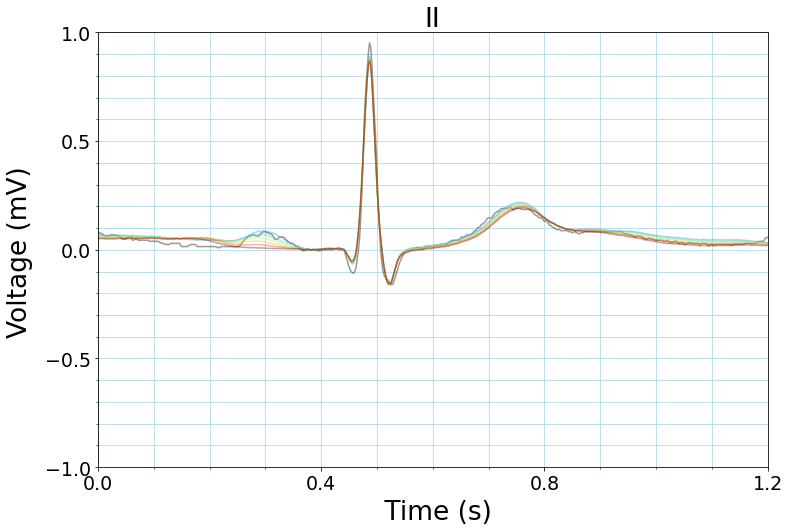}
        \hfill
        (g)\includegraphics[width=0.45\textwidth]{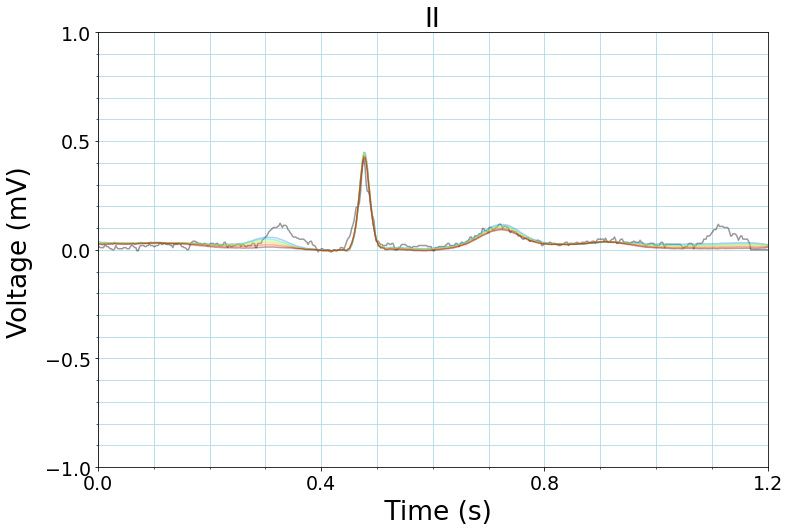}
        \hspace{0.6cm}
        (h)\includegraphics[width=0.45\textwidth]{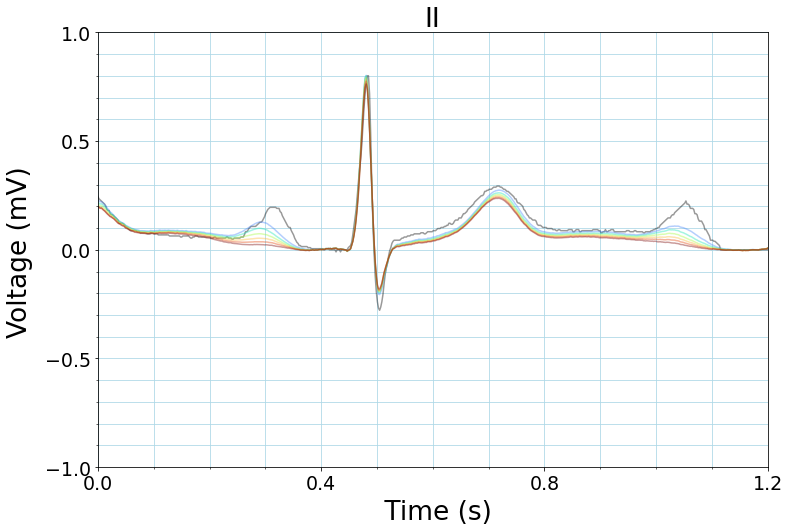}
        \hfill
        \label{fig:af_8_success}
      \caption{(a) to (h) show lead II latent traversals using our \qLST system on sinus rhythm (SR) samples where \qLST successfully introduces atrial fibrillation (AF) according to the classifier. The examples show \qLST introduces AF in lead II by removing the P wave and leaves other features of the ECG in tact.}
    \end{minipage}
\end{figure*}

\qLST can increase the posterior class probability for arbitrary input/class combinations. Figure \ref{fig:af_8_success} shows examples where \qLST introduces atrial fibrillation (AF) to sinus rhythm ECGs.  

\subsection{Failure to introduce class}

\begin{figure*}[h]
    \vspace{-0.3cm}
  \centering
    \begin{minipage}[h]{\textwidth}
        \centering\includegraphics[width=0.9\textwidth]{images/legend.png}
    \end{minipage}
    \begin{minipage}[h]{\textwidth}
        (a)\includegraphics[width=0.45\textwidth]{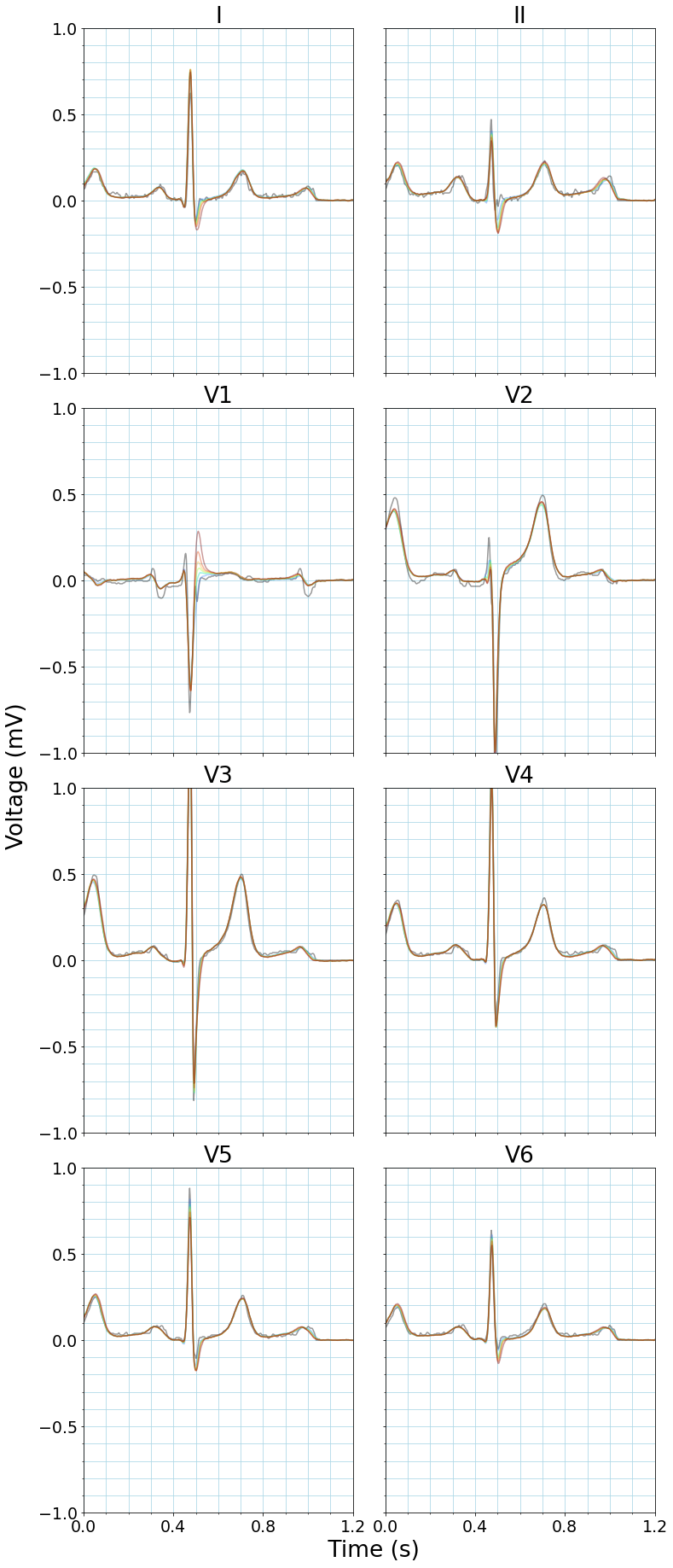}
        (b)\includegraphics[width=0.45\textwidth]{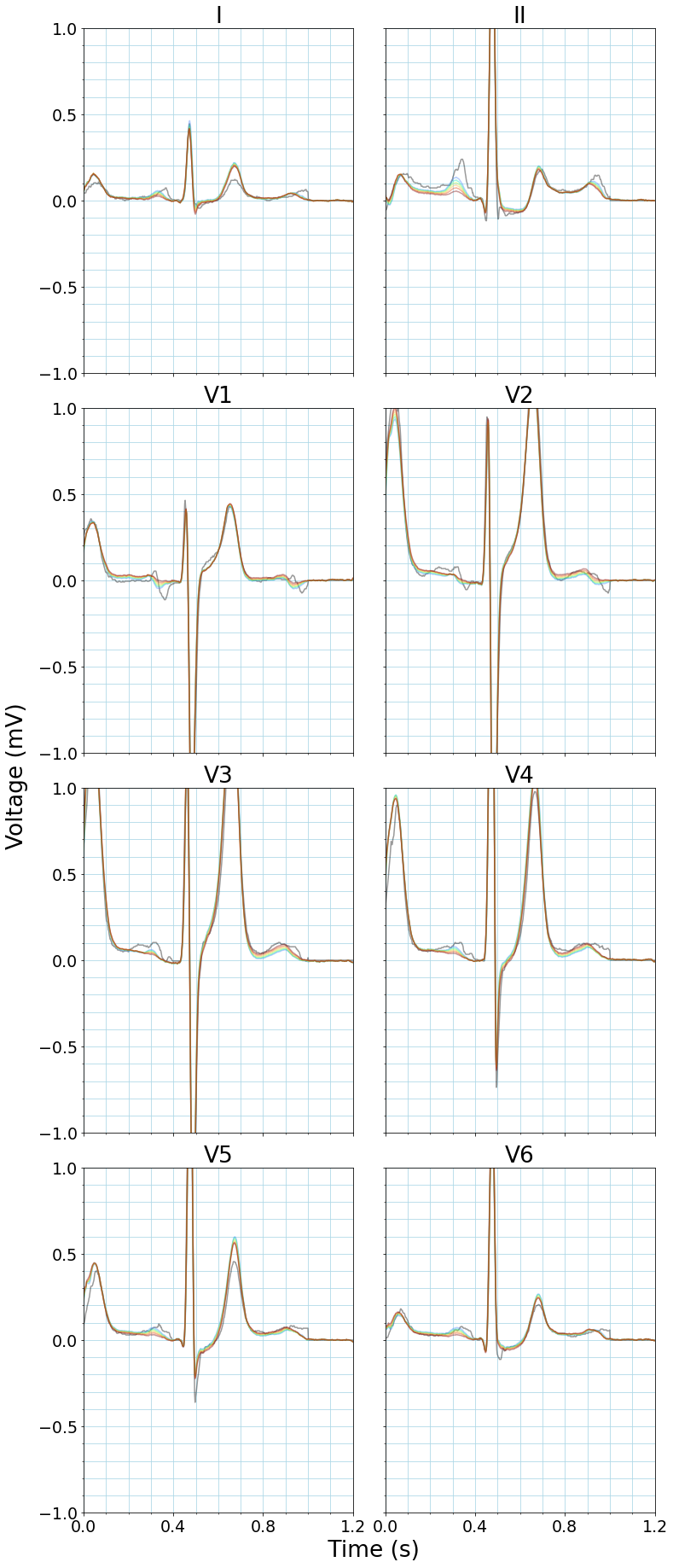}
       \caption{\textbf{(a)} Latent traversal showing our \qLST system applied to a failed classification of a right bundle branch block (RBBB) sample. Through our method we show that even tough a physician deemed this sample to be a RBBB, the classifier expects a higher S wave in V1. This is a known feature of RBBB and not visibly present in the original ECG, explaining the misclassification. \textbf{(b)} 8 lead latent traversal for an atrial fibrillation (AF) ECG. In this example the \qLST system failed to increase the AF classification probability above 0.4. The system attempts to reduce the P-wave (which is missing in AF patients) but fails to remove it completely. This is likely due to the large P-wave in the original ECG and the limited size of changes made by the \qLST system.}
    \label{fig:rbtb_af}
    \end{minipage}
\end{figure*}

\qLST is not able to fully increase class probabilities for all ECGs. Figure \ref{fig:rbtb_af}(b) and \ref{fig:af_8_failed} show examples where \qLST is unable to introduce atrial fibrillation (AF) to an ECG. 

\subsection{Other classifier results}
\begin{figure*}[h]
    \centering
    \includegraphics[width= \textwidth, height=6cm]{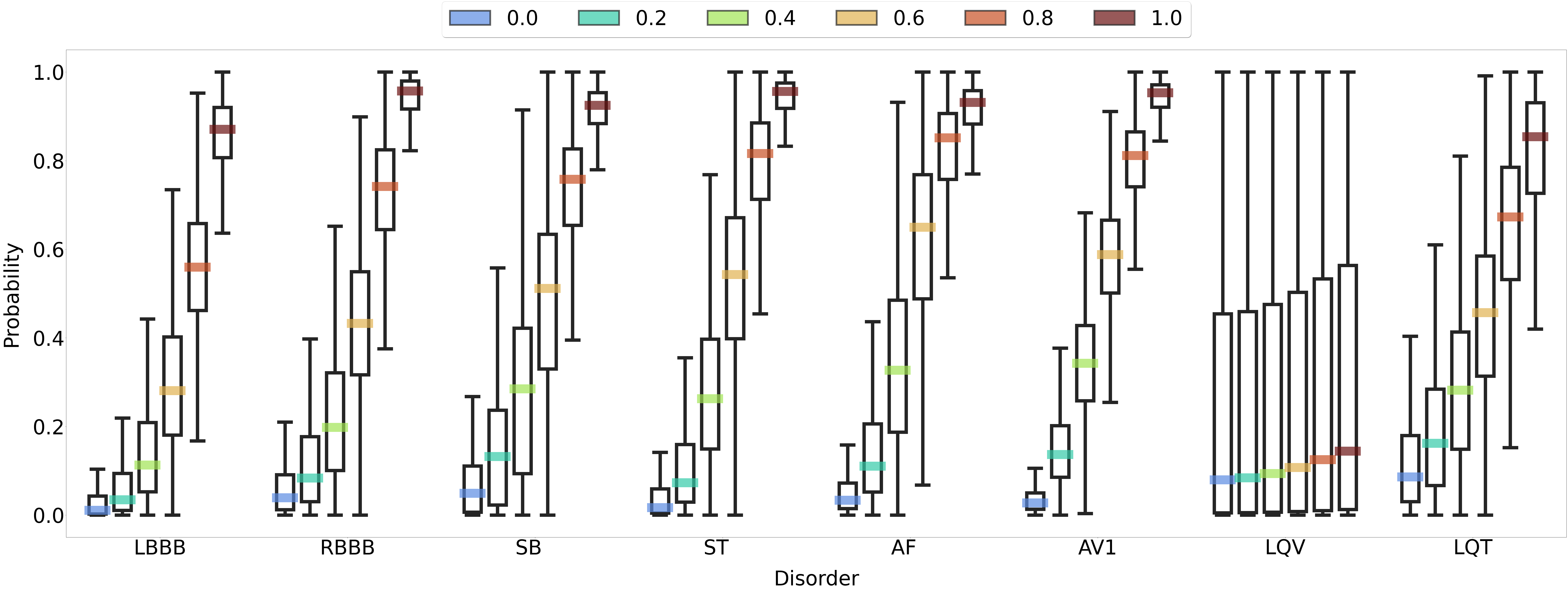}
    \caption{Boxplot of class probability for query values 0, 0.2, 0.4, 0.6, 0.8 and 1. Each class is grouped on the x axis, class probability is shown on the y-axis. The \qLST results shown are created using the MLP classifier architecture. The plot shows that the \qLST system is not able to increase the low QRS voltage (LQV) class probability for this classifier. We suspect this is due to poor performance of the MLP classifier on this class.}
    \label{fig:MLP_boxplot}
\end{figure*}

\begin{figure*}[h]
    \centering
    \includegraphics[width= \textwidth, height=6cm]{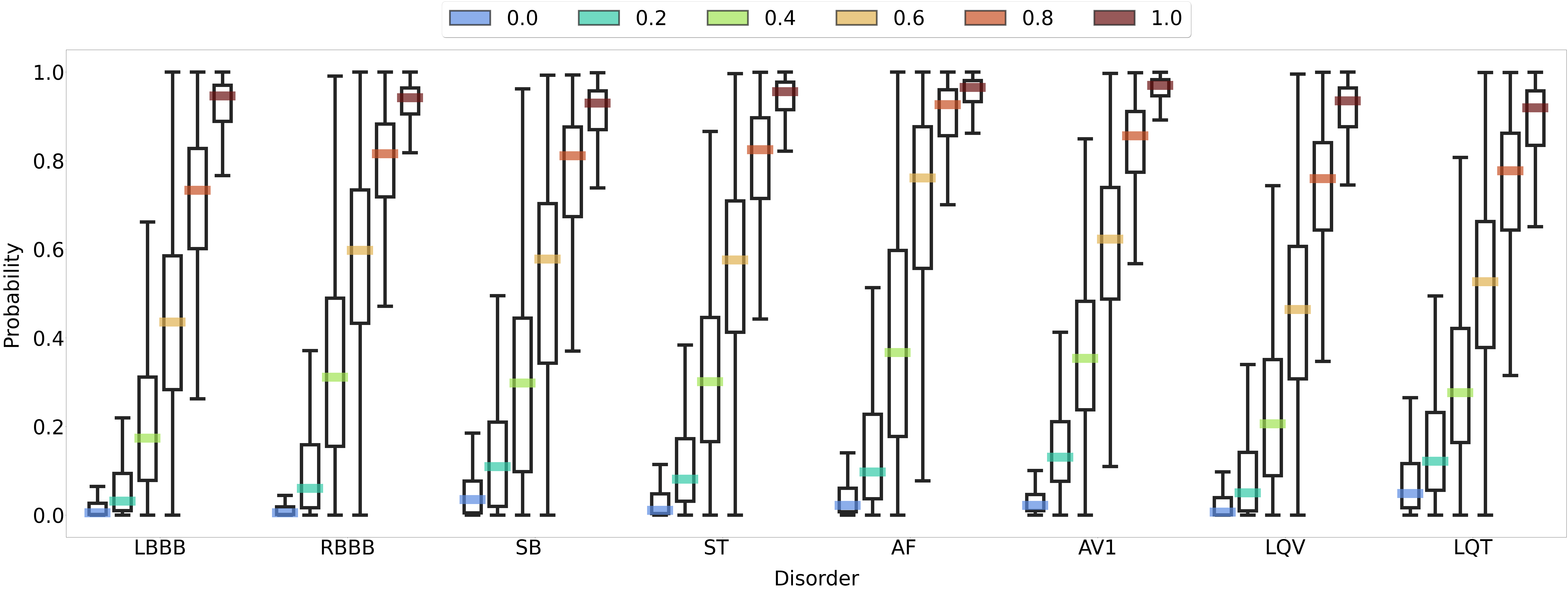}
    \caption{Boxplot of class probability for query values 0, 0.2, 0.4, 0.6, 0.8 and 1. Each class is grouped on the x axis, class probability is shown on the y-axis. The \qLST results shown are created using the classifier architecture proposed by \cite{Hannun2019}. The plot shows that \qLST preforms similar to ECGResNet (Figure \ref{fig:ECGnet_boxplot}) for this classifier and how \qLST is able to approximate class probabilities for the given queries.}
    \label{fig:Hanun_boxplot}
\end{figure*}

\begin{figure*}[h]
    \vspace{-1cm}
  \centering
    \begin{minipage}[h]{\textwidth}
        \centering\includegraphics[width=0.9\textwidth]{images/legend.png}
    \end{minipage}
    \begin{minipage}[h]{\textwidth}
        (a)\includegraphics[width=0.45\textwidth]{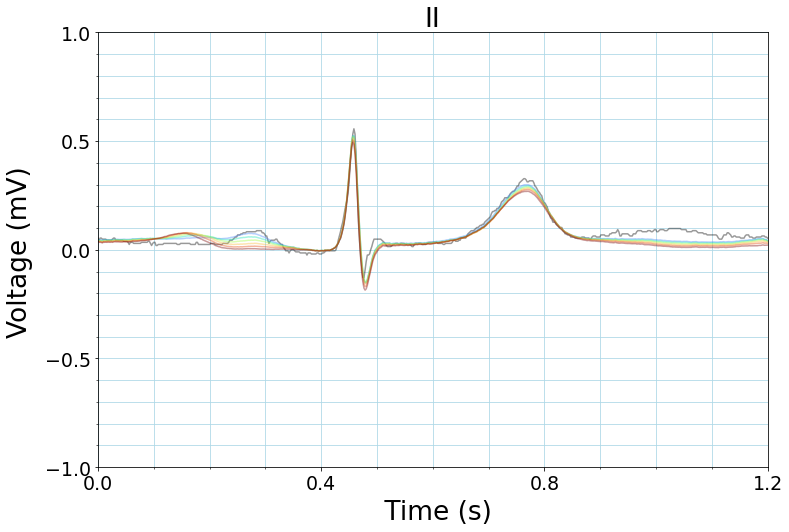}
        \hfill
        (b)\includegraphics[width=0.45\textwidth]{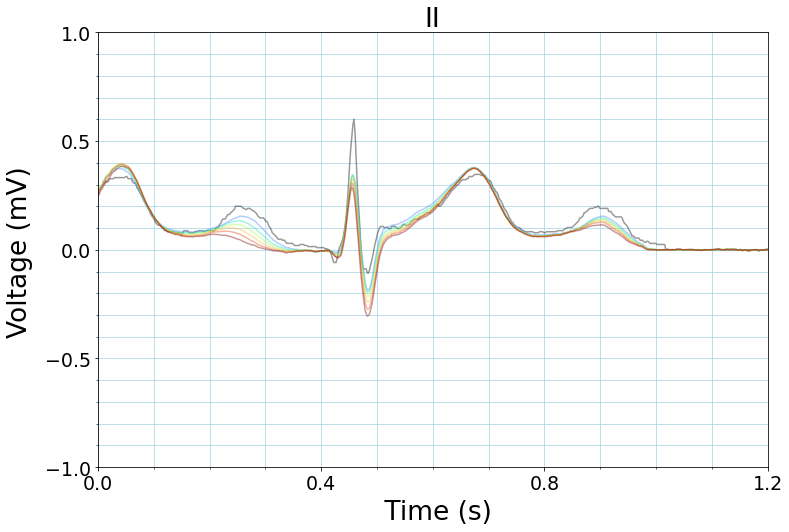}
        \hfill
        (c)\includegraphics[width=0.45\textwidth]{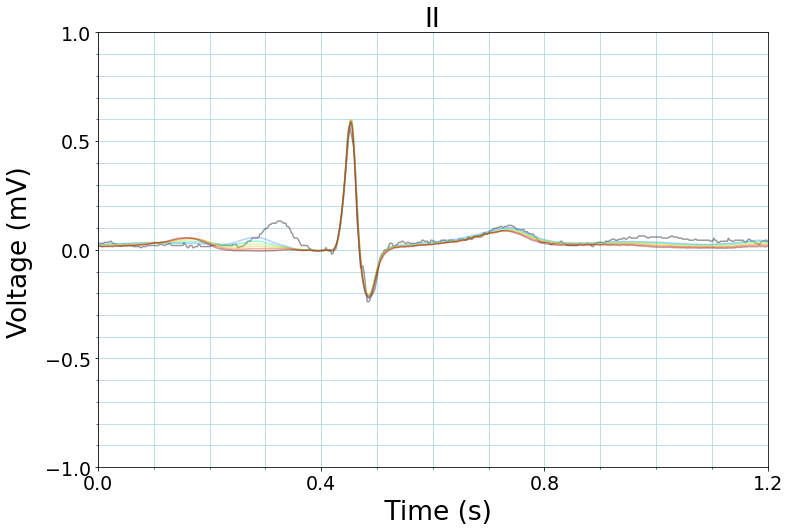}
        \hfill
        (d)\includegraphics[width=0.45\textwidth]{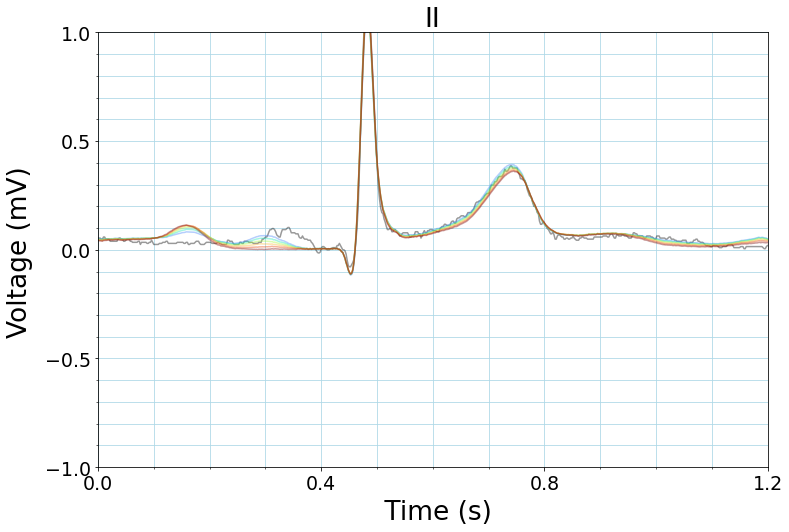}
        \hfill
        (e)\includegraphics[width=0.45\textwidth]{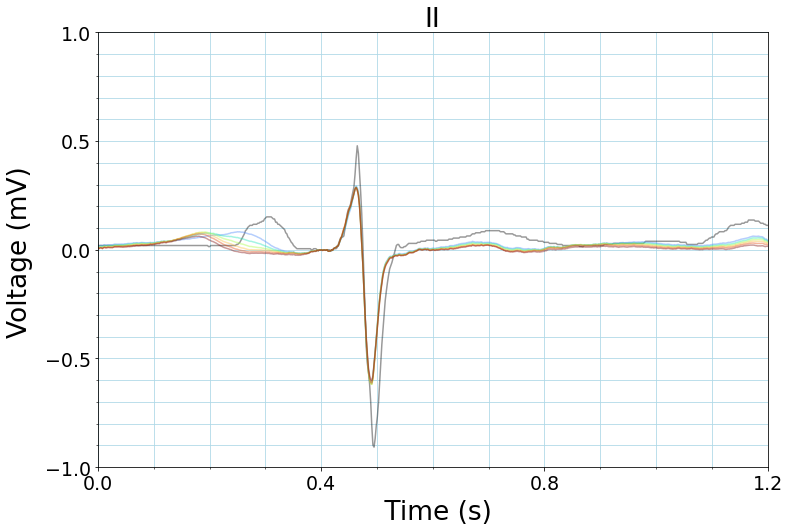}
        \hfill
        (f)\includegraphics[width=0.45\textwidth]{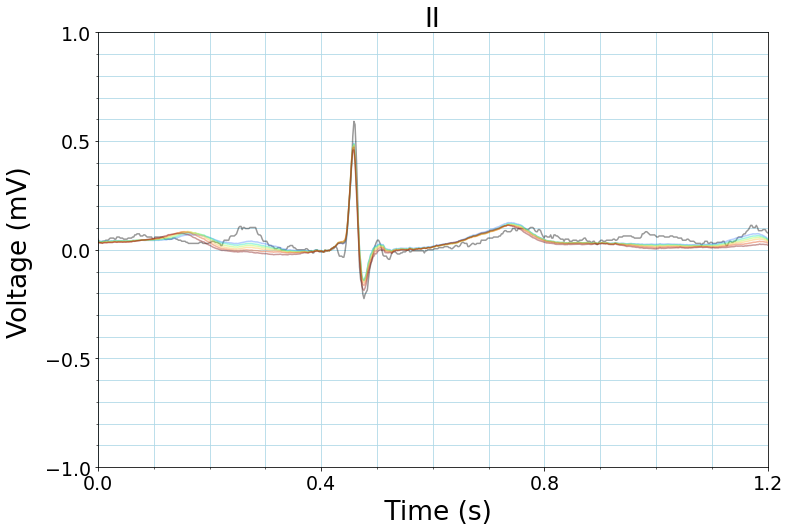}
        \hfill
        (g)\includegraphics[width=0.45\textwidth]{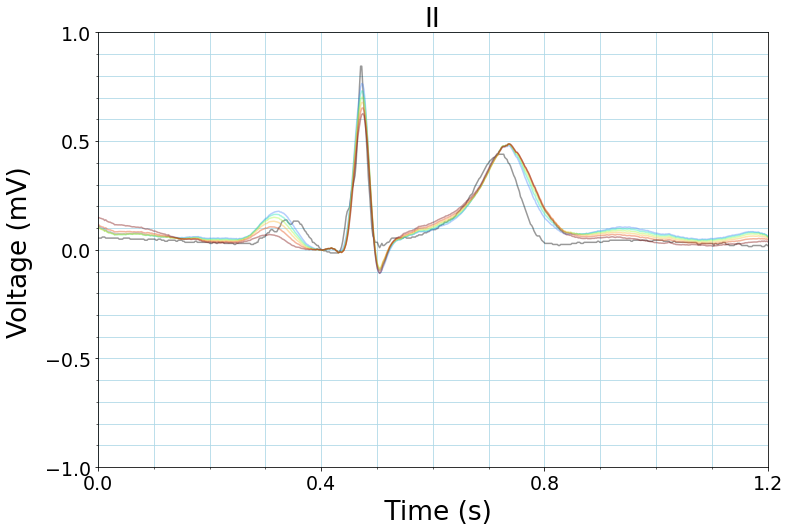}
        \hspace{0.6cm}
        (h)\includegraphics[width=0.45\textwidth]{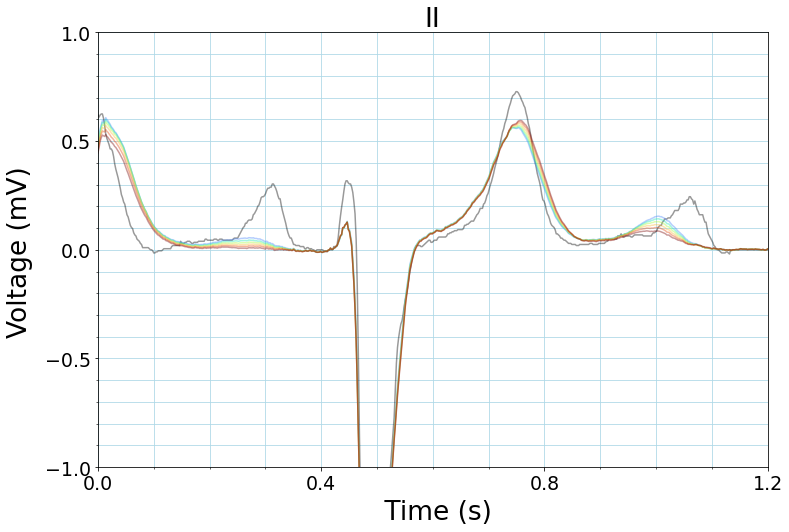}
        \hfill
        \label{fig:af_8_failed}
      \caption{(a) to (h) show lead II latent traversals using our \qLST system on sinus rhythm (SR) samples where \qLST fails to introduce atrial fibrillation (AF) according to the classifier. Figures \ref{fig:ECGnet_boxplot}, \ref{fig:MLP_boxplot} and \ref{fig:Hanun_boxplot} show that these failures are rare. For all examples the failure appears related to a failure to remove the P wave. Note that in example (h) the P wave of the next wave is still present on the right hand side of the ECG.}
    \end{minipage}
\end{figure*}

Figure \ref{fig:MLP_boxplot} shows the results as a boxplot for MLP based classifier. The plot shows that the \qLST system is not able to increase the low QRS voltage (LQV) class probability. Figure \ref{fig:Hanun_boxplot} shows the results as a boxplot for a classifier proposed by \cite{Hannun2019}.

\subsection{Implementation details}

\begin{figure}[H]
    \centering\includegraphics[width=0.28\textwidth]{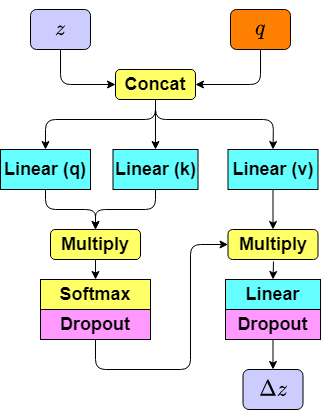}
  \caption{Schematic of the attention module used in \qLST}
  \label{fig:attention}
\end{figure}

All results presented in this paper were created using \qLST models trained using the same set of hyper-parameters. The training regime consisted of 3 steps: (1) Train the classifier of interest; (2) Train a VAE model in a semi-supervised fashion (e.g. with class-size based reweighing); (3) Train \qLST models for each class, using the pretrained classifier and VAE with fixed weights. During training \qLST only had access to the latent representations of ECG samples, the input queries (sampled from a uniform distribution) and the output of the classifier for the class being investigated. The value of $\alpha$ in equation \ref{eq:loss} was set to $25(1 - \phi)$ where $\phi$ is the absolute difference between the original classification of the sample $\hat{y}$ and the query value $q$. The \qLST network itself consisted of a single multi-headed attention module with 5 attention heads where dropout was applied to the attention mask and final output layer with probability $p=0.1$. A schematic of the used attention module is shown in Figure \ref{fig:attention}. The VAE network was trained on the entire dataset consisting of 864,051 ECGs, for these ECGs the physician annotated labels (where available) and the automatically generated MUSE labels were used for class-size based reweighing. The classifiers and the \qLST networks were trained on a subset of 337662 ECGs where physician annotated labels were available. 

\end{document}